\pdfoutput=1

\documentclass[11pt]{article}

\usepackage[final]{acl}

\usepackage{times}
\usepackage{latexsym}

\usepackage[T1]{fontenc}

\usepackage[utf8]{inputenc}

\usepackage{microtype}

\usepackage{inconsolata}

\usepackage{graphicx}

\usepackage{booktabs}
\usepackage{hyperref}
\usepackage{multirow}

\usepackage[most]{tcolorbox}
\newtcolorbox{promptbox}{
colback=gray!5,  %
colframe=black!75, %
left=1em, %
right=1em, %
top=1em, %
bottom=1em, %
sharp corners, %
boxrule=1pt %
}

\usepackage{soul}  %
\usepackage{xcolor}  %

\definecolor{lightblue}{RGB}{212, 235, 255}
\colorlet{lightblue}{lightblue!150}
\definecolor{lightorange}{RGB}{255, 204, 168}
\colorlet{lightorange}{lightorange!70}
\definecolor{blueblue}{RGB}{79,123,200}

\newcommand{\colorlightblue}[1]{\sethlcolor{lightblue}\hl{#1}}

\newcommand{\colorlightorange}[1]{\sethlcolor{lightorange}\hl{#1}}

\colorlet{lightblueblue}{lightblue!400}
\newcommand{\blue}[1]{\textcolor{blueblue}{{#1}}}

\definecolor{promptcolor}{RGB}{0,127,255}

\title{Can Large Language Models Identify Authorship?}

\author{Baixiang Huang$^1$ \ \ \ Canyu Chen$^1$ \ \ \ Kai Shu$^2$\thanks{Corresponding author} \\
$^1$Illinois Institute of Technology \ \ \ $^2$Emory University \\
\small \texttt{\{bhuang15, cchen151\}@hawk.iit.edu} \ \ \ \texttt{kai.shu@emory.edu}
}

\newcommand{\RQOne}{Can LLMs perform zero-shot, end-to-end authorship verification effectively?}
\newcommand{\RQTwo}{Are LLMs capable of accurately attributing authorship among multiple candidates authors (e.g., 10 and 20)?}
\newcommand{\RQThree}{Can LLMs provide explainability in authorship analysis, particularly through the role of linguistic features?}

\begin{document}
\maketitle
\begin{abstract}
The ability to accurately identify authorship is crucial for verifying content authenticity and mitigating misinformation. Large Language Models (LLMs) have demonstrated an exceptional capacity for reasoning and problem-solving. However, their potential in authorship analysis remains under-explored. Traditional studies have depended on hand-crafted stylistic features, whereas state-of-the-art approaches leverage text embeddings from pre-trained language models. These methods, which typically require fine-tuning on labeled data, often suffer from performance degradation in cross-domain applications and provide limited explainability. This work seeks to address three research questions: (1) \RQOne{} (2) \RQTwo{} (3) \RQThree{} Moreover, we investigate the integration of explicit linguistic features to guide LLMs in their reasoning processes. Our assessment demonstrates LLMs' proficiency in both tasks without the need for domain-specific fine-tuning, providing explanations into their decision making via a detailed analysis of linguistic features. This establishes a new benchmark for future research on LLM-based authorship analysis\footnote{Code and data are publicly available at~\href{https://llm-authorship.github.io/\#canllm-identify-authorship}{https://llm-authorship.github.io}}.
\end{abstract}

\section{Introduction}

Authorship analysis is the study of writing styles to determine the authorship of a piece of text, impacting areas from forensic investigation, such as distinguishing between murders and suicides \cite{chaski2005forensic_court}, to tracking terrorist threats \cite{winter2019terrorism,cafiero2023aa_qanon}. It addresses challenges in digital forensics and cybersecurity, including the fight against misinformation, impersonation, and cyber threats such as phishing and deceptive social media posts \cite{argamon2018forensic,shu2020misinfo,stiff2022detect_generated_misinfo}. Authorship analysis is essential for tracing cyber threats to their sources, combating plagiarism to uphold intellectual property rights \cite{stamatatos2011plagiarism}, and identifying compromised accounts \cite{barbon2017social_media}. In addition, it helps link user accounts across social platforms \cite{shu2017link_user,sinnott2021social_mdeia} and detect fraudulent activities such as fake reviews \cite{ott2011data_opinion_spam}.

Historically, authorship analysis relied on methods based on human expertise to distinguish between authors \cite{mosteller1963aa_bayes}. Later, a line of research known as stylometry emerged, which developed various features to quantify writing styles \cite{holmes1994survey_aa}. The evolution continued with the adoption of rule-based computational linguistic methods \cite{stamatatos2009survey}. The development of statistical algorithms provides the capability to handle data with higher dimensions, enabling more expressive representations. These methods relied heavily on extensive text preprocessing and feature engineering \cite{bozkurt2007aa, seroussi2014data_imdb_aa_topic_model}. 

Compared to traditional statistical methods, deep learning techniques require less feature engineering. Among these techniques, pre-trained language models (PTMs) are widely used for representing authorship \cite{huang2024aa_llm}. These models, built predominantly on BERT-based architectures \cite{devlin2018bert} and contrastive learning paradigms, demonstrate efficacy in domain-specific applications. However, they fall short in cross-domain scenarios \cite{rivera2021luar}. The performance of these methods also declines significantly with shorter query texts \cite{eder2015aa_text_size,grieve2019aa_data_size} and limited data from the candidate authors. This reduction in performance limits their applicability in real-world situations, where data scarcity and diversity are the norms. While some studies have attempted to overcome these challenges by applying text style transfer to learn content-independent style representations, they have not addressed the cross-domain issue effectively \cite{boenninghoff2019av,wegmann2022av_transfer}. These deep learning methods require extensive time and labeled data for training, are not effectively applicable across different data domains, and suffer from limited explainability.

Despite the rapid development of LLMs, there has been insufficient analysis and evaluation of their capabilities in authorship analysis \cite{huang2024aa_llm}. Some initial studies have utilized GPT-3 \cite{brown2020gpt3} for annotating data \cite{patel2023lisa} before employing a T5 Encoder \cite{raffel2020t5} for learning representations of authorship. LLMs have demonstrated proficiency in zero-shot learning scenarios within domains lacking extensive resources \cite{kojima2022zero_shot_reasoners}. However, their ability to grasp subtle nuances of language and extract critical features for authorship identification has not been extensively examined. Consequently, this paper aims to investigate the potential of LLMs for authorship identification by addressing the following research questions:

\begin{itemize}
    \item \textbf{RQ1:} \RQOne{}
    \item \textbf{RQ2:} \RQTwo{}
    \item \textbf{RQ3:} \RQThree{}
\end{itemize}

\begin{figure*}[h]
    \centering
    \includegraphics[width=1\textwidth]{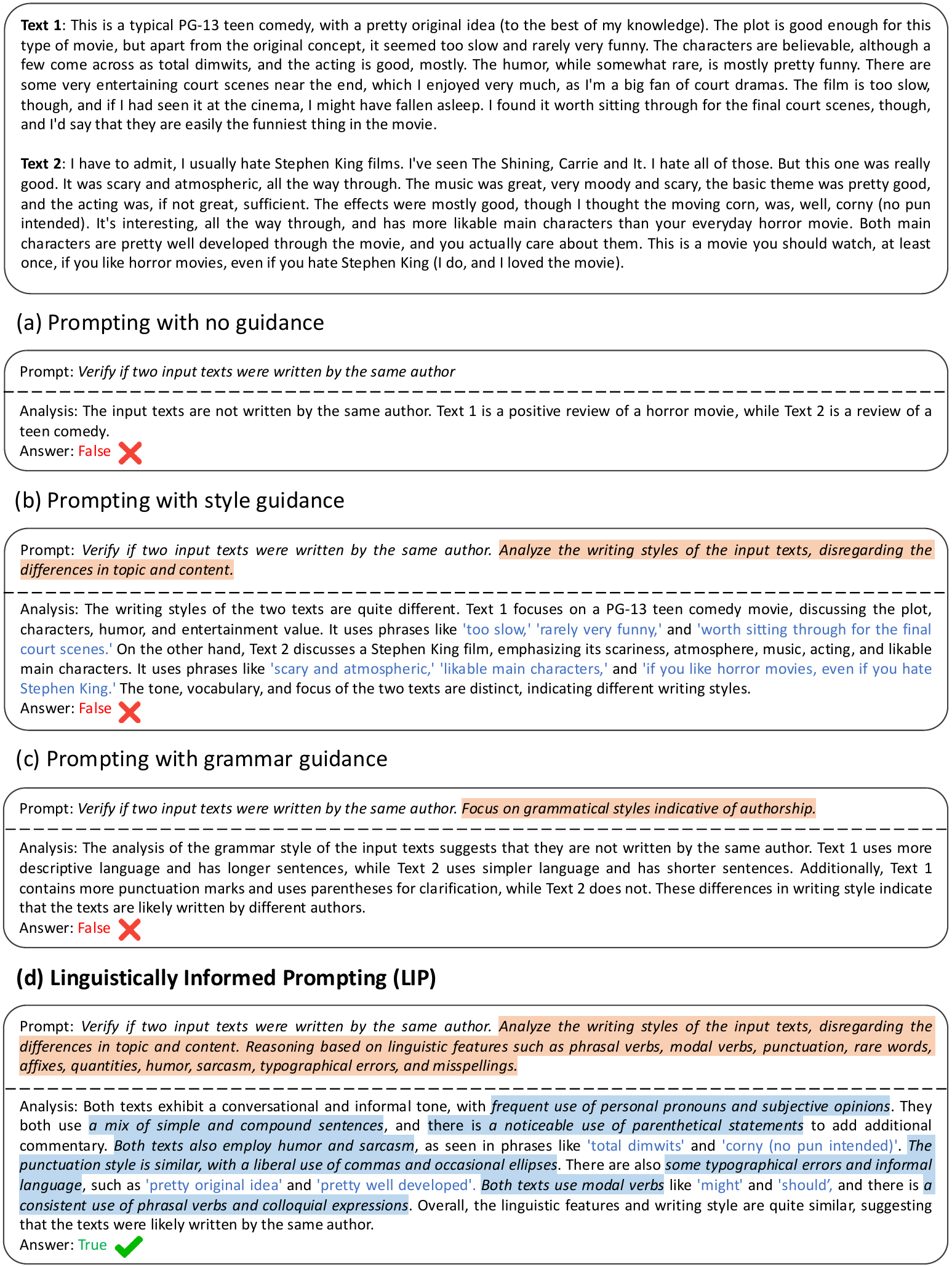}
    \caption{\textbf{A Comparison Between Linguistically Informed Prompting (LIP) and other Prompting Strategies for Authorship Verification}. ``Analysis'' and ``Answer'' are the output of prompting GPT-4. Only LIP strategy correctly identifies that the given two texts belong to the same author. Text colored in \colorlightorange{orange} highlights the differences compared to vanilla prompting with no guidance. Text colored in \colorlightblue{blue} indicates the linguistically informed reasoning process. \blue{Blue} text represents the text referenced from the original documents. 
    }
    \label{fig:case-study}
\end{figure*}

We also propose a prompting technique named Linguistically Informed Prompting (LIP) to guide LLMs in identifying linguistic features that are used in practice by forensic linguists \cite{grant2022linguistic}. This approach exploits the inherent linguistic knowledge embedded within LLMs, unleashing their potential to discern subtle stylistic nuances and linguistic patterns indicative of individual authorship. Figure \ref{fig:case-study} demonstrates the application of the LIP method in verifying authorship through linguistic feature analysis using GPT-4. It compares two texts from the Blog dataset \cite{schler2006data_blog}. Analysis can provide specific linguistic evidence such as the use of informal language, punctuation patterns, and typographical errors.

Our empirical evaluation includes data with different genres and topics to validate the robustness and versatility of LLMs. The results demonstrate that LLMs can effectively perform zero-shot authorship verification and attribution, thereby obviating the need for fine-tuning. With the introduction of linguistic guidance, LLMs are further leveraged for authorship analysis, where our LIP technique sets a new benchmark for LLM-based authorship prediction. The key contributions of this work are summarized as follows:

\begin{itemize}
    \item We conduct a comprehensive evaluation of LLMs in authorship attribution and verification tasks. Our results demonstrate that LLMs outperform existing BERT-based models in a zero-shot setting, showcasing their inherent stylometric knowledge essential for distinguishing authorship. This enables them to excel in authorship attribution and verification across low-resource domains without the need for domain-specific fine-tuning.

    \item We develop a pipeline for authorship analysis with LLMs, encompassing dataset preparation, baseline implementation, and evaluation. Our novel Linguistically Informed Prompting (LIP) technique guides LLMs to leverage linguistic features for accurate authorship analysis, enhancing their reasoning capabilities.

    \item Our end-to-end approach improves the explainability of authorship analysis. This approach elucidates the reasoning and evidence behind authorship predictions, shedding light on how various linguistic features influence these predictions. This contributes to a deeper understanding of the mechanisms behind LLM-based authorship identification.
\end{itemize}

\section{Datasets}

We choose two representative datasets to highlight the importance of user-generated content such as emails and social media posts. The first dataset is the Enron Email dataset, which consists of approximately half a million messages from senior Enron managers. This dataset offers insights into corporate communication, featuring long texts and a high variance in text-per-author \cite{klimt2004data_enron}. The second dataset is the Blog Authorship Attribution corpus \citet{schler2006data_blog}. This corpus includes blog articles from 2004 and earlier, collected from Blogger.com, comprising over 680,000 

\clearpage
\newpage
\clearpage
\newpage
\noindent
posts from more than 19,000 authors, averaging 35 posts per author. The texts in this dataset are relatively short, with an average length of 79 tokens for the top five authors. For data preprocessing, we remove duplicate texts and authors with fewer than two texts and filter out non-English texts.

If the data formulation is not balanced, we observe that LLMs tend to predict any two given texts are written by different authors. LLMs are trained on varied datasets with many authors, which makes them better at detecting differences in writing style than similarities. The lack of multiple works by the same author in training data may prevents LLMs from learning individual authorial nuances. For our experiments, we organize and sub-sample these datasets for authorship attribution and authorship verification tasks separately. For the authorship verification task, we ensure a balanced distribution of positive and negative cases, meaning half of the texts are from the same author, and the other half are written by different authors. We sample 30 pairs of texts and conduct experiments three times for each dataset. We also make sure that all the authors and texts are unique. For the authorship attribution task, we ensure that the classes and authors are balanced: in each of the three repetitions, every query text is written by a different author, and all candidate authors are unique with only one correct author who also wrote the query text. We randomly select 10 or 20 different texts from different authors for every repetition. All texts in our sampled subsets are unique. We have saved the sampled data and used the same subsets across all baselines. The preprocessing code and sampled data are available on our GitHub repository.

\begin{figure}[t]
    \centering
    \includegraphics[width=0.99\linewidth]{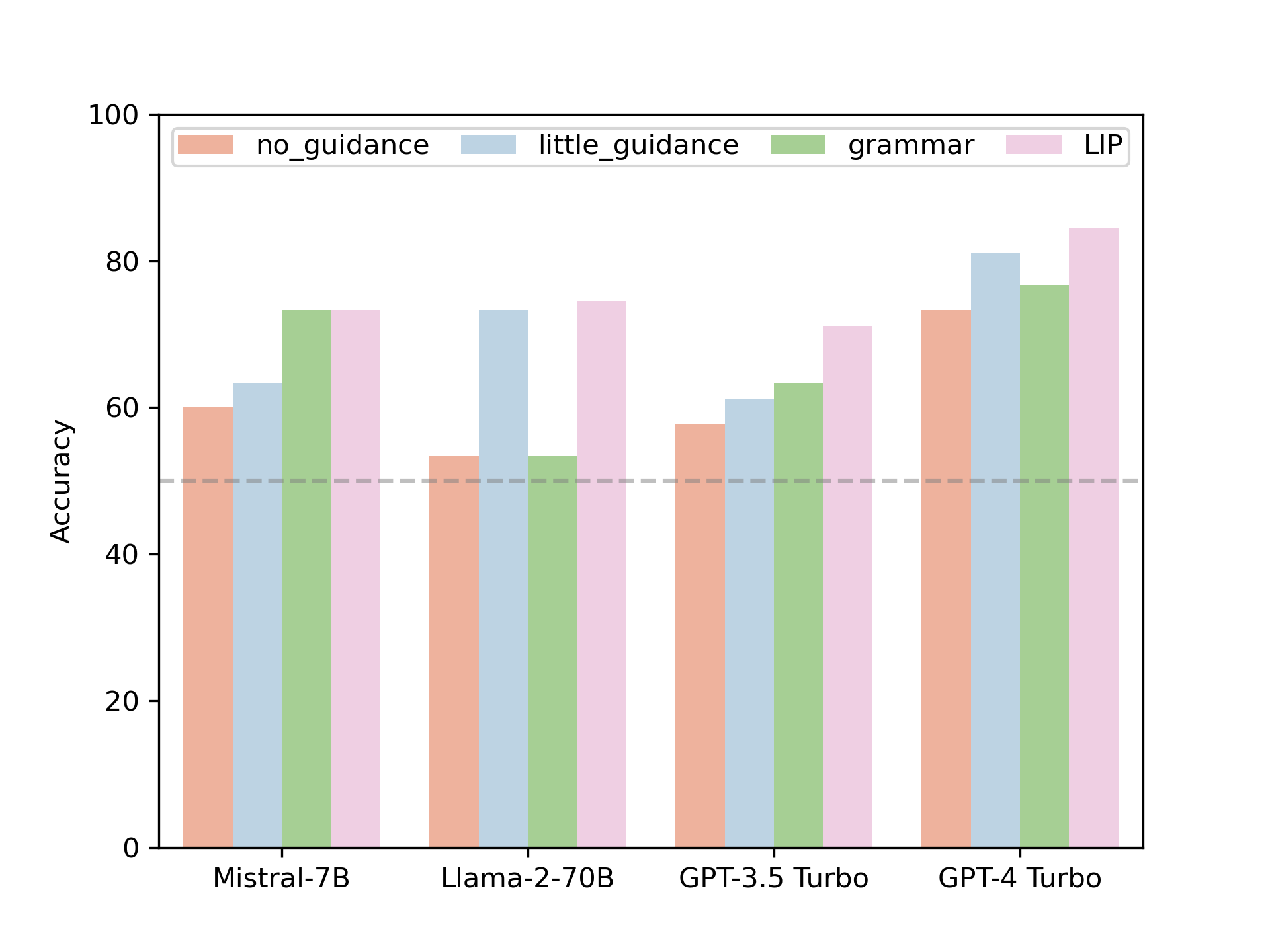}
    \caption{Authorship verification results in terms Accuracy (\%) on the Blog dataset.}
    \label{fig:verify-blog}
\end{figure}

\begin{figure}[h]
    \centering
    \includegraphics[width=0.99\linewidth]{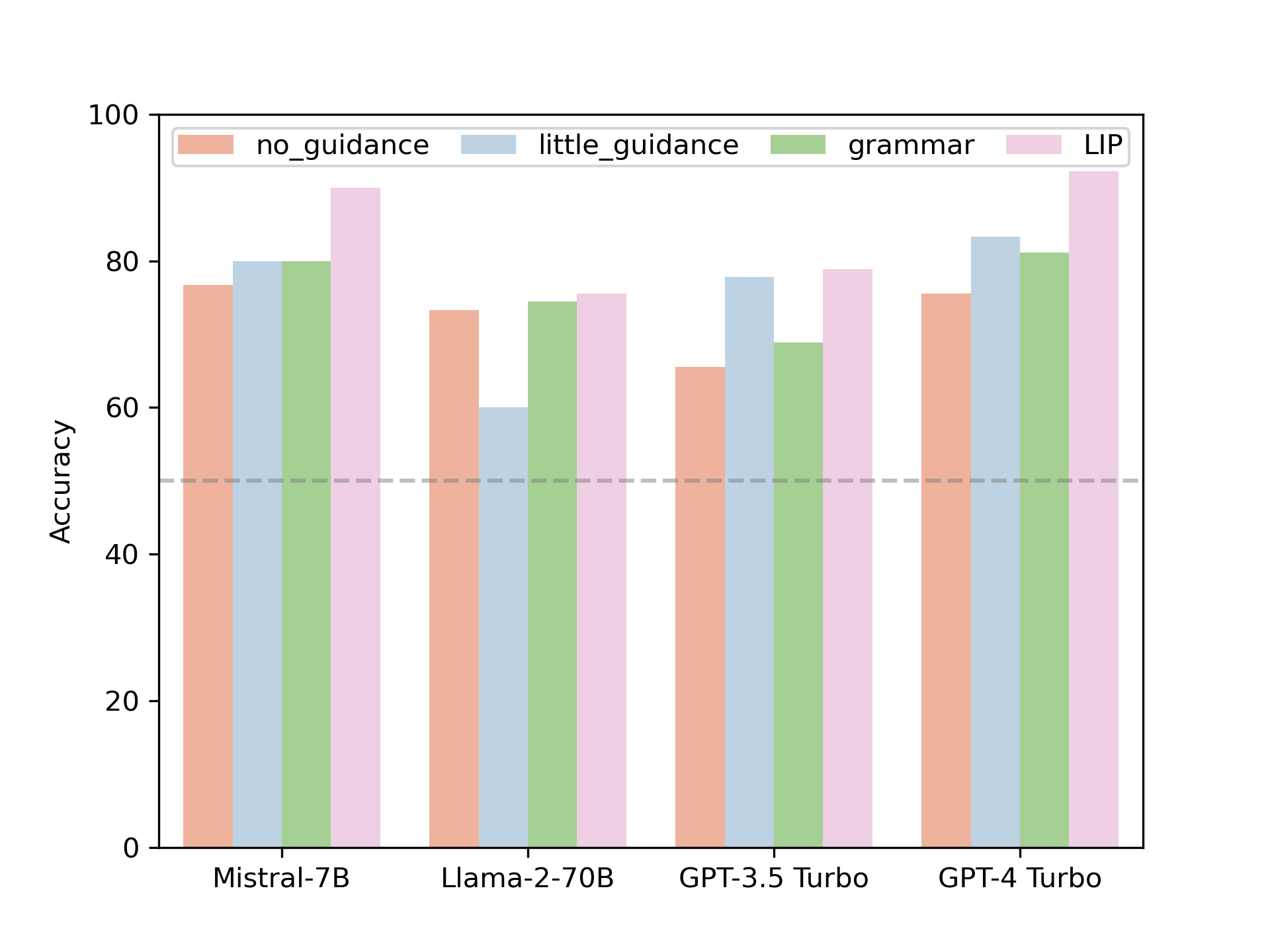}
    \caption{Authorship verification results in terms Accuracy (\%) on the Email dataset.}
    \label{fig:verify-email}
\end{figure}

\section{Authorship Verification (RQ1)}
Authorship verification involves assessing whether a single candidate author is the correct author of a query text. This process can be formulated as a one-class classification problem \cite{koppel2007av}. An important variant of authorship verification is to predict whether two or more given pieces of text were written by the same author or not \cite{koppel2012av_two_doc}. The complexity of the author verification problem is amplified when it involves only a pair of documents for comparison. This scenario substantially limits the amount of reference material available for analysis. In this work, we address the problem of determining if two texts were authored by the same author.

The difference between "{LIP}", "\texttt{no\_guidance}", "\texttt{little\_guidance}", and "\texttt{grammar\_guidance}", lies in their specificity and focus. "\texttt{no\_guidance}" is the prompt that only offers a basic task description. "\texttt{little\_guidance}" narrows this down by emphasizing writing style over content or topic differences, while "\texttt{grammar\_guidance}" further specifies a focus on grammatical style that reflects authorship. "{LIP}" provides a specific list of linguistic features to analyze, such as phrasal verbs, modal verbs, punctuation, and typographical errors, ensuring a comprehensive and focused analysis. Such linguistic guidance allows for a more nuanced assessment, particularly effective in cases where an author writes on varied topics across different domain, thereby making "{LIP}" superior in authorship verification. Figure \ref{fig:prompt_verify} in Appendix \ref{sec:appendix-prompt} shows the prompts we used for this task.

\begin{table*}[t]
\centering
\resizebox{0.87\linewidth}{!}{
\begin{tabular}{l|l|ccc|ccc} 
\toprule
\multicolumn{1}{l}{}           &                  & \multicolumn{3}{c|}{10 candidate authors}        & \multicolumn{3}{c}{20 candidate authors}          \\
\midrule
Model                          & Prompt           & Weighted F1    & Macro F1       & Micro F1       & Weighted F1    & Macro F1       & Micro F1        \\
\midrule
TF-IDF                         &                  & 11.89          & 11.89          & 20.00          & 1.20           & 1.20           & 5.00            \\
BERT                           &                  & 42.22          & 42.22          & 50.00          & 27.50          & 27.50          & 33.33           \\
RoBERTa                        &                  & 34.44          & 34.44          & 43.33          & 23.33          & 23.33          & 28.33           \\
ELECTRA                        &                  & 34.67          & 34.67          & 40.00          & 10.55          & 10.55          & 13.33           \\
DeBERTa                        &                  & 38.89          & 38.89          & 46.67          & 19.09          & 19.09          & 23.33           \\
\midrule
\multirow{4}{*}{GPT-3.5 Turbo} & no\_guidance     & 16.67          & 15.15          & 20.00          & 24.50          & 18.69          & 27.50           \\
                               & little\_guidance & 27.22          & 24.75          & 33.33          & 37.83          & 27.94          & 37.50           \\
                               & grammar          & 31.85          & 31.85          & 40.00          & 29.50          & 23.06          & 32.50           \\
                               & LIP              & 30.56          & 27.78          & 40.00          & 33.33          & 25.20          & 37.50           \\
\midrule                               
\multirow{4}{*}{GPT-4 Turbo}   & no\_guidance     & 36.67          & 33.33          & 36.67          & 37.50          & 32.20          & 35.00           \\
                               & little\_guidance & 36.67          & 33.33          & 36.67          & 40.83          & 37.12          & 40.00           \\
                               & grammar          & 58.89          & 53.54          & 60.00          & 59.84          & 49.86          & 57.50           \\
                               & \textbf{LIP}     & \textbf{84.45} & \textbf{79.40} & \textbf{86.67} & \textbf{60.50} & \textbf{55.00} & \textbf{62.50} \\
\bottomrule                               
\end{tabular}
}
\caption{Authorship attribution results with 10 and 20 candidate authors in terms of Weighted F1(\%), Macro F1(\%), and Micro F1(\%) on the Blog dataset.}
\label{table:exp-blog}
\end{table*}

\begin{table*}[t]
\centering
\resizebox{0.87\linewidth}{!}{
\begin{tabular}{l|l|ccc|ccc} 
\toprule
\multicolumn{1}{l}{}           &                  & \multicolumn{3}{c|}{10 candidate authors} & \multicolumn{3}{c}{20 candidate authors}  \\ 
\midrule
Model                          & Prompt           & Weighted F1 & Macro F1 & Micro F1         & Weighted F1 & Macro F1 & Micro F1         \\ 
\midrule
TF-IDF                         &                  & 15.25       & 15.25    & 26.67            & 5.95        & 5.95     & 8.33             \\
BERT                           &                  & 41.67       & 41.67    & 46.67            & 38.89       & 38.89    & 45.00            \\
RoBERTa                        &                  & 36.67       & 36.67    & 43.33            & 22.06       & 22.06    & 26.67            \\
ELECTRA                        &                  & 28.89       & 28.89    & 36.67            & 22.78       & 22.78    & 30.00            \\
DeBERTa                        &                  & 27.22       & 27.22    & 33.33            & 21.67       & 21.67    & 25.00            \\ 
\midrule
\multirow{4}{*}{GPT-3.5 Turbo} & no\_guidance     & 23.33       & 21.21    & 23.33            & 20.00       & 16.92    & 20.00            \\
                               & little\_guidance & 28.89       & 26.26    & 30.00            & 26.67       & 24.24    & 26.67            \\
                               & grammar          & 18.89       & 17.17    & 20.00            & 20.00       & 17.93    & 20.00            \\
                               & LIP              & 22.22       & 20.20    & 23.33            & 28.89       & 26.26    & 30.00            \\ 
\midrule
\multirow{4}{*}{GPT-4 Turbo}   & no\_guidance     & 73.33       & 66.67    & 73.33            & 67.22       & 57.66    & 70.00            \\
                               & little\_guidance & 71.67       & 65.15    & 73.33            & 73.33       & 66.67    & 73.33            \\
                               & grammar          & 80.00       & 77.98    & 83.33            & 73.89       & 67.76    & 76.67            \\
                               & \textbf{LIP}     & \textbf{88.89}       & \textbf{86.87}    & \textbf{90.00}            & \textbf{77.22}       & \textbf{73.33}    & \textbf{80.00}            \\
\bottomrule
\end{tabular}
}
\caption{Authorship attribution results with 10 and 20 candidate authors in terms of Weighted F1(\%), Macro F1(\%), and Micro F1(\%) on the Email dataset.}
\label{table:exp-email}
\end{table*}

Comparing the performance of LLMs with the BERT model in our zero-shot setting is not straightforward. This is because we instruct LLMs to directly output their final answer, as illustrated in the system instructions from Figure \ref{fig:prompt_verify} in Appendix. In contrast, traditional pre-trained models, such as BERT, require fine-tuning along with a prediction head (typically a trained machine learning classifier) to map hidden embeddings to the final output. Our experiments indicate that employing cosine similarity with BERT embeddings, which are predominantly distributed around 0.9, makes it challenging to distinguish authorship for verification purposes in this zero-shot setting. The experimental results for both TF-IDF and BERT are provided in Appendix \ref{sec:appendix-add-res}.

Evaluation metrics for this task include accuracy, precision, recall, and F1 Score. Accuracy is a fundamental metric that measures the proportion of correct predictions out of the total predictions made. Precision focuses on the proportion of true positive predictions within the pool of positive predictions, evaluating the model's ability to avoid false positives. Recall assesses the model's ability to identify all actual positives, reflecting its capability to minimize false negatives. The F1 Score harmonizes precision and recall by providing a single metric that balances both aspects through the calculation of their harmonic mean.

The experiment results are demonstrated in Figure \ref{fig:verify-blog} and \ref{fig:verify-email}. They provide a comparative analysis of the performance of LLMs in authorship verification tasks across two different datasets. Four models are evaluated with four different prompt settings. GPT-4 Turbo consistently outperforms the other models in both datasets, indicating its superior capability in understanding authorship. The LIP method generally yields the highest scores across all metrics for most models. Across both datasets, performance metrics improve as the level of prompt guidance increases from no guidance to LIP. This trend underscores the importance of linguistic guidance in leveraging LLMs for authorship verification. The analysis of these figures reveals that the effectiveness of LLMs in authorship verification tasks can be significantly influenced by the type of prompt guidance provided.

\section{Authorship Attribution (RQ2)}
In this section, we present a comprehensive analysis of experiments conducted to evaluate the efficacy of our proposed models on the zero-shot authorship attribution task. We selected the Blog and Enron email datasets to ensure a robust assessment across different domains and genres. Figure \ref{fig:prompt_aa} in Appendix \ref{sec:appendix-prompt} shows the prompts of this task, we utilize four prompt similar to the authorship verification task, with LIP being the most effective due to its linguistic guidance effect. The experiments were structured to compare the performance of LLMs not only against each other but also against established benchmarks in the field, such as TF-IDF and BERT-based models. 

Authorship attribution, the task of determining the most likely author of a given text from a set of candidates, is commonly formulated as a multi-class, single-label text classification problem. Tables \ref{table:exp-blog} and \ref{table:exp-email} provide a overview of the performance of various models. These models were evaluated across two different datasets (Blog and Email) with varying numbers of candidate authors (10 and 20).

Weighted F1, micro F1, and macro F1 are used as evaluation metrics. Weighted F1 gives an average F1 score weighted by class size. Micro F1 calculates the overall average F1 score, combining all classes, and is sensitive to class imbalance. Macro F1 computes the unweighted average of F1 scores across classes, treating each class equally, ideal for assessing minority class performance.

\begin{table}[!t]
\centering
\resizebox{0.99\linewidth}{!}{
\begin{tabular}{llccc}
\toprule
Dataset                & Prompt           & Weighted F1 & Macro F1 & Micro F1  \\
\midrule
\multirow{4}{*}{Blog}  & no\_guidance     & 10.00       & 9.09     & 13.33     \\
                       & little\_guidance & 6.89        & 6.26     & 10.00     \\
                       & grammar          & 7.22        & 6.57     & 10.00     \\
                       & LIP              & 10.56       & 9.90     & 13.33     \\
\midrule                       
\multirow{4}{*}{Email} & no\_guidance     & 22.22       & 20.20    & 26.67     \\
                       & little\_guidance & 22.45       & 20.40    & 26.67     \\
                       & grammar          & 15.00       & 13.64    & 20.00     \\
                       & LIP              & 29.44       & 28.53    & 33.33    \\
\bottomrule                     
\end{tabular}
}
\caption{Mistral's performance on the authorship attribution task with 10 candidate authors.}
\label{table:aa-mistral}
\end{table}

\begin{table}[!t]
\centering
\resizebox{0.99\linewidth}{!}{
\begin{tabular}{llccc}
\toprule
Model                           & Prompt               & Weighted F1 & Macro F1 & Micro F1  \\
\midrule
\multirow{10}{*}{GPT-3.5} & phrasal verbs        & 22.67       & 20.61    & 30.00     \\
                                & modal verbs          & 20.95       & 20.04    & 26.67     \\
                                & punctuation          & 23.06       & 22.60    & 33.33     \\
                                & rare words           & 26.00       & 23.64    & 33.33     \\
                                & affixes              & 23.00       & 20.91    & 30.00     \\
                                & quantities           & 19.44       & 18.23    & 30.00     \\
                                & humor                & 23.22       & 21.66    & 33.33     \\
                                & sarcasm              & 23.89       & 21.72    & 33.33     \\
                                & typos & 24.67       & 23.36    & 33.33     \\
                                & misspellings         & 28.33       & 26.67    & 40.00     \\
\midrule                                
\multirow{10}{*}{GPT-4}   & phrasal verbs        & 62.22       & 56.57    & 63.33     \\
                                & modal verbs          & 56.67       & 51.52    & 56.67     \\
                                & punctuation          & 71.11       & 67.27    & 73.33     \\
                                & rare words           & 62.22       & 56.57    & 63.33     \\
                                & affixes              & 75.56       & 71.32    & 76.67     \\
                                & quantities           & 75.56       & 71.32    & 76.67     \\
                                & humor                & 66.67       & 60.61    & 70.00     \\
                                & sarcasm              & 72.22       & 65.66    & 73.33     \\
                                & typos & 55.56       & 50.51    & 56.67     \\
                                & misspellings         & 46.67       & 42.42    & 46.67    \\
\bottomrule                                
\end{tabular}
}
\caption{Ablation study on the impact of 10 linguistic features for the Blog dataset (with 10  candidate authors).}
\label{table:ablation}
\end{table}

We also tested Llama 2 and Mistral 7B. However, input texts from the datasets we used for evaluating other LLMs are too long and exceed the context limit of Llama 2 because of their context length limitations (4k tokens for Llama 2 and 8k for Mistral, versus 16k for GPT-3.5 Turbo and 128k for GPT-4 Turbo). Therefore the experiment on Mistral are shown in a separate table (Table \ref{table:aa-mistral}).

\begin{figure*}[!t]
    \centering
    \includegraphics[width=1\linewidth]{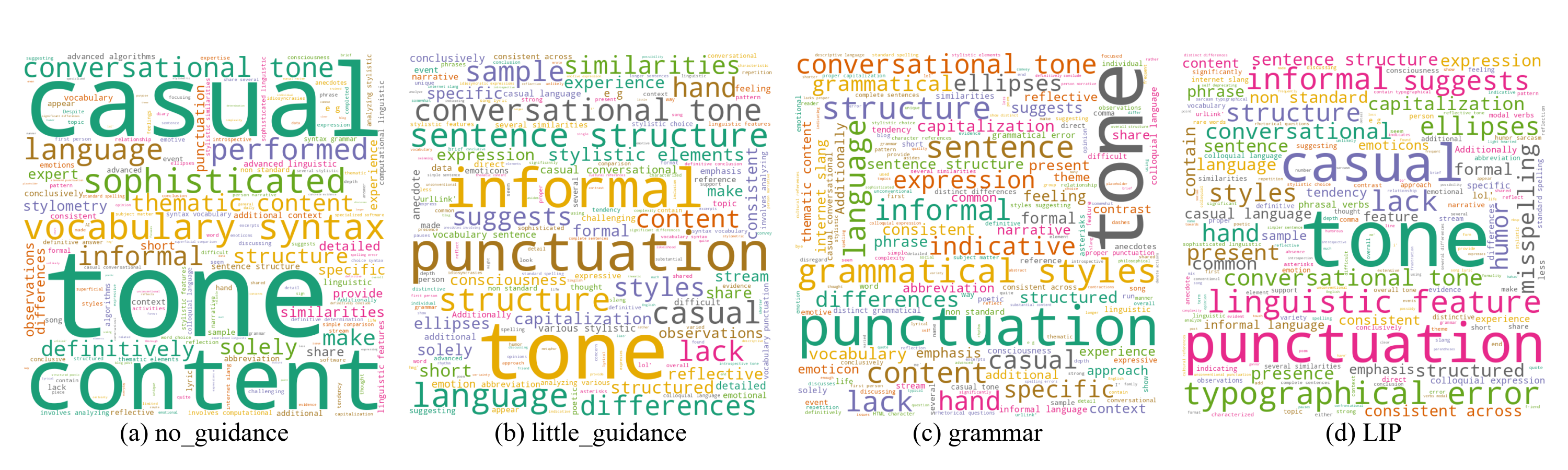}
    \caption{Visualization of GPT-4's Authorship Verification Analysis on the Blog dataset with four Guidance Levels.}
    \label{fig:word-cloud-blog-verify}
    \captionsetup{belowskip=-10pt}
\end{figure*}

\begin{figure*}[!t]
    \centering
    \includegraphics[width=1\linewidth]{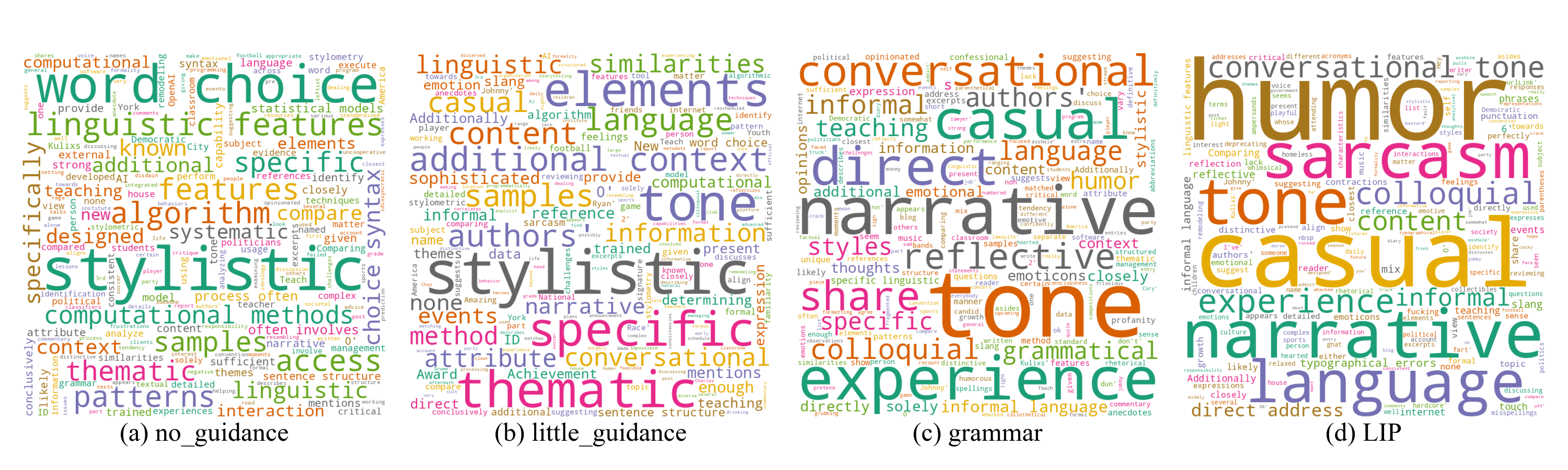}
    \caption{Visualization of GPT-4's Authorship Attribution Analysis on the Blog dataset with four Guidance Levels.}
    \label{fig:word-cloud-blog-aa}
\end{figure*}

Table \ref{table:ablation} outlines the results of an ablation study focused on evaluating the impact of various linguistic features on the performance of LLMs. This study examines how the exclusion of specific features affects the models' abilities. Features such as affixes and quantities are crucial for GPT-4 Turbo, while misspellings hold more significance for GPT-3.5 Turbo. We use all of these linguistic features provided by forensic linguistics \cite{grant2022linguistic}, and find that LLMs perform optimally when all of these features are provided, as shown in the LIP technique in Table \ref{table:exp-blog}, allowing LLMs to determine which features to utilize.

The experiment results highlight the superiority of GPT-4 Turbo over BERT-based language models and basic statistical approaches, such as TF-IDF. These advancements not only demonstrate significant improvements in scores but also show robustness against increased task complexity. The incorporation of linguistic guidance for LLMs markedly improves performance and generates more explainable authorship analysis. This progression emphasizes the importance of adopting LLMs for complex authorship analysis tasks. Key observations are summarized as follows:

\begin{itemize}
    \item Across both datasets and both sets of candidate authors, GPT-4 consistently outperforms traditional models and those using GPT-3.5 in all metrics. Specifically, linguistic-guided GPT-4 approaches show remarkable improvements, indicating the effectiveness of these advanced models in understanding and attributing authorship with higher precision. There is a clear trend of performance improvement when transitioning from traditional models and BERT-based models to LLMs. This suggests that more recent and sophisticated models are significantly better at handling the complexities of authorship attribution.
    
    \item The increase in the number of candidate authors from 10 to 20 generally results in a decline in performance across all models. However, the decline is less pronounced for LLMs, suggesting that they is better equipped to handle increased complexity and maintain higher levels of performance.
    
    \item Traditional pre-trained language models such as BERT, RoBERTa, DeBERTa, and ELECTRA exhibit considerable variability in their performance, suggesting they require domain-specific fine-tuning and need to be combined with other classifiers to perform this task effectively. TF-IDF shows the lowest performance across all metrics in both datasets.
\end{itemize}

\vspace{-0.2cm}
\section{Explainability (RQ3)}
LLMs can provide explanations in natural language regarding their decision-making processes and provide linguistic evidence that references the original text, which is invaluable for verifying and understanding their conclusions. Our LIP method improves upon this by generating meaningful analyses of writing styles, which are instrumental in distinguishing between authors. We have made all the input texts and the corresponding authorship analyses, as produced by LLMs, accessible in our GitHub repository.

Figure \ref{fig:word-cloud-blog-verify} displays word clouds that represent the outputs from GPT-4 during the authorship verification task. These visualizations demonstrate the model's focus and specificity in analysis, where the density and size of terms indicate their prominence in the output. As guidance increases from none to LIP, there is a clear shift from general and diverse terms to more specific linguistic features. The word cloud for LIP, being the most effective, underscores a thorough analysis by highlighting particular linguistic characteristics.

Similarly, the word clouds in Figure \ref{fig:word-cloud-blog-aa} illustrate that LLMs can offer in-depth explanations for authorship attribution tasks. The effectiveness and focus of these explanations can be significantly improved through explicit linguistic guidance, which directs the model to base its decisions on linguistic attributes used in practice \cite{grant2022linguistic}. The word cloud of the LIP method prominently features terms such as "humor", "sarcasm", "casual", and "colloquial." This demonstrates that with LIP, the LLM is steered towards making decisions grounded in linguistic features, especially high-level and complex features such as humor and sarcasm. The specificity achieved through the LIP method highlights the model's ability to provide clear and focused explanations for its authorship decisions, offering a notable improvement over traditional methods that rely on hidden embeddings. The enhanced clarity in the LLM's outputs not only facilitates a better understanding of the decision-making process but also has the potential to increase the reliability of the authorship analysis process.

\section{Related Work}
In this section, we review the literature on traditional and contemporary methods of authorship analysis, as well as research on utilizing LLMs for authorship analysis and related tasks.

\subsection{Authorship Analysis}
The primary goal of authorship analysis is to analyze writing styles to determine authorship. It encompasses two main tasks: authorship attribution and verification. Authorship attribution, also known as authorship identification, aims to attribute a previously unseen text of unknown authorship to one of a set of known authors. Authorship verification involves determining whether a single candidate author wrote the query text by comparing text similarities \cite{koppel2007av}. This process requires establishing whether a query text was written by a specific author, compared to a set of their known works. Authorship attribution can be broken down into a series of authorship verification instances, focusing on measuring text similarity based on stylistic features. We specifically focus on closed-set authorship attribution, which deals with a predetermined, finite list of potential authors that always includes the true author of a query text. Authorship verification can also be seen as a specific case of authorship attribution, but with only one potential author.

Central to these tasks is the extraction of useful authorship features from textual data using natural language processing methods such as n-grams \cite{sharma2018aa_ngram}, POS-tags \cite{sundararajan2018aa_pos}, topic modeling \cite{seroussi2014data_imdb_aa_topic_model}, and Linguistic Inquiry and Word Count (LIWC) \cite{uchendu2020aa_neural_liwc}. More recently, the focus has shifted towards extracting embeddings from text, considering both content and style while often disregarding external contextual cues. These embeddings, serving as a numeric representation of a text segment, facilitate further analysis. When comparing a document embedding with another from the same author, the representation tends to orient toward the author's style rather than the document's content \cite{huertas2022part}.

\citet{barlas2020aa_cross_domain_bert} found that BERT models perform well when dealing with large vocabularies, outperforming multi-headed RNNs. \citet{fabien2020aa_bert_classifiy} fine-tuned a BERT model for authorship attribution. They showed that incorporating stylometric and hybrid features into an ensemble model enhances its performance. \citet{huertas2022part} introduced a semi-supervised contrastive learning approach using a BERT-based model for cross-domain authorship attribution and profiling. \citet{rivera2021luar} also explored cross-domain authorship representation learning through contrastive learning, revealing that neural authorship representations learned by deep learning models, such as Sentence-BERT (SBERT), are not universal. They concluded that topic diversity and the size of the training dataset are crucial for effective zero-shot cross-domain transfer. For instance, models trained on the Reddit comments \cite{baumgartner2020data_reddit_pushshift} exhibited significantly better transfer than those trained on the Amazon Reviews corpus \cite{ni2019data_amazon} and the Fanfiction dataset \cite{bevendorff2020data_pan20}. Deep learning methods, despite their potential, require substantial training time and labeled data, offer limited generalization capabilities, and lack explainability. In contrast, our approach, which leverages the intrinsic linguistic knowledge and zero-shot reasoning abilities of LLMs, does not require fine-tuning and is effective in low-resource domains.

\subsection{Large Language Models}
Large Language Models (LLMs) excel at text generation, achieving a level of fluency and coherence that closely mimics human writing. Hence, numerous studies have focused on differentiating LLM-generated text from human-written text using various machine learning methods \cite{huang2024aa_llm,uchendu2020aa_neural_liwc,tang2023llm_text,wu2023survey_llm_text,yang2023survey_llm_text}. In comparison, our research evaluates LLMs' capabilities in authorship verification and attribution, which are complex reasoning tasks. Unlike pre-trained language models (PTMs) that often require specific fine-tuning for different tasks, LLMs have an inherent capacity for reasoning and problem-solving. This is leveraged through instruction-based few-shot or zero-shot learning, allowing them to effectively conduct reasoning tasks with minimal examples \cite{brown2020gpt3,kojima2022zero_shot_reasoners}. 

The application of LLMs in authorship analysis, particularly in authorship attribution and authorship verification, is rarely explored. Traditional methods have primarily used LLMs for auxiliary tasks, such as data extraction and annotation, rather than fully utilizing their capabilities \cite{patel2023lisa}. In contrast, our work is pioneering in exploring LLMs' end-to-end potential for authorship analysis tasks. This not only demonstrates the versatility and effectiveness of LLMs in complex linguistic tasks but also sets a new benchmark for future research in the field.

Moreover, this novel application of LLMs in authorship analysis aims to overcome the limitations of traditional methods, such as extensive feature engineering. Unlike BERT-based models, which require computationally expensive fine-tuning and large amounts of domain-specific data for optimal performance \cite{grieve2019aa_data_size}, LLMs can generalize across various domains without any fine-tuning, addressing the issue of domain specificity \cite{barlas2020aa_cross_domain_bert}. They are also capable of handling shorter texts, reducing the need for long inputs to derive meaningful representations \cite{eder2015aa_text_size}. A key advantage of our LLM-based approach is its ability to provide understandable natural language explanations for its predictions, addressing the lack of transparency in traditional models' hidden text embeddings \cite{rivera2021luar}. This improvement in explainability and versatility represents a significant advancement in overcoming the challenges related to data, domain specificity, text length requirements, and explainability faced by earlier methods.

\section{Conclusion}
This paper explores how to leverage LLMs for authorship analysis. Through comprehensive evaluation, it demonstrates that LLMs, equipped with the novel Linguistically Informed Prompting (LIP) technique, excel at identifying authorship without the need for domain-specific fine-tuning. By directly applying our end-to-end methods to authorship attribution and verification tasks, we aim to bypass the intermediate steps of feature extraction and manual annotation. This approach not only surpasses traditional and state-of-the-art methods in performance, especially in zero-shot and low-resource settings, but also enhances the explainability of authorship predictions by illuminating the role of linguistic features. The findings underscore the potential of LLMs to revolutionize authorship analysis, offering robust solutions for digital forensics, cybersecurity, and combating misinformation. This work paves the way for future research and applications in LLM-based authorship prediction.

\section{Limitations}
\paragraph{Scalability with Increasing Number of Authors} The effectiveness of the method when the number of candidate authors increases is a major limitation. In real-world scenarios, especially in contexts like social media and large forums, the number of potential authors can be vast. If the model's performance degrades with more candidates, this restricts its utility in broader applications. Another potential limitation is the evaluation of machine-generated text for authorship analysis, particularly as machine-generated content becomes more common and sophisticated. Our method may not effectively distinguish between human-authored and machine-generated texts.

\paragraph{Explainability} Although authorship analysis by LLMs offers a level of explainability through the linguistic features or insights highlighted during the analysis, the mechanistic interpretability of how these decisions are made at the neuronal level within the LLMs is not explored. This means that while we can observe the decisions that are made, the fundamental neural activities and interactions that lead to these decisions remain a black box. This lack of deeper explainability can be a drawback, particularly in critical applications where understanding the precise reasoning process is necessary for trust and verification.

\section{Ethics Statement}
The potential to reveal the identities of anonymous authors presents an ethical challenge. The paper discusses applications such as linking user accounts across platforms and identifying compromised accounts. These applications raise privacy concerns and ethical questions about surveillance and the profiling of individuals based on their writing style. The use of such methods must be carefully managed to protect individual privacy and adhere to ethical standards, particularly in sensitive areas such as journalism, political dissent, or corporate whistleblowing. Ensuring that authorship attribution methods are not used to undermine privacy rights or expose individuals to risks without their consent is crucial.

\section*{Acknowledgments}
We thank our anonymous reviewers for their constructive feedback and recommendations. This material is based upon work supported by the U.S. Department of Homeland Security under Grant Award Number 17STQAC00001-07-04, NSF awards (SaTC-2241068, IIS-2339198, and POSE-2346158), a Cisco Research Award, and a Microsoft Accelerate Foundation Models Research Award. The views and conclusions contained in this document are those of the authors and should not be interpreted as necessarily representing the official policies, either expressed or implied, of the U.S. Department of Homeland Security and the National Science Foundation.

\bibliography{custom}

\clearpage
\newpage
\appendix

\section{Impact Statement}
Authorship verification and attribution play an essential role in various applications such as combating misinformation \citep{shu2020misinfo,hanley2024misinfo,chen2024can,chen2024combatingmisinformation,chen2022combating,huang2024edit_hallu,chen2024canediting,stiff2022disinfo,beigi2024model}, protecting intellectual property rights \citep{meyer2007plagiarism,stamatatos2011plagiarism}, identifying fraudulent activities \citep{ott2011data_opinion_spam,afroz2012hoax_fraud}, tracking terrorist threats \citep{winter2019terrorism,cafiero2023aa_qanon}, and aiding general criminal investigations \citep{koppel2008law_enforcement,argamon2018forensic,belvisi2020forensic}. 

\section{Future Work}

The advent of LLMs has complicated the problem of authorship attribution since it is increasingly challenging to distinguish between LLM-generated and human-written texts~\cite{huang2024aa_llm}. The hardness of differentiating the content produced by humans and machines potentially undermines the integrity of authorship, threatens the credibility of digital content and endangers safety of online space~\citep{solaiman2023evaluating,vidgen2024introducing}. More effort is desired to protect human authorship from the threat of LLM-generated content.

\section{Experiment Setup}
The baselines used in this paper include: TF-IDF, pre-trained language models like BERT (\texttt{bert-base-uncased}) \cite{devlin2018bert}, RoBERTa (\texttt{roberta-base}) \cite{liu2019roberta}, DeBERTa (\texttt{deberta-base}) \cite{he2020deberta}, and ELECTRA (\texttt{electra-base-discriminator}) \cite{clark2020electra}, alongside LLMs represented by GPT-3.5 Turbo (\texttt{1106-preview}) and GPT-4 Turbo (\texttt{1106-preview}).
We use {GPT-3.5 Turbo (\texttt{1106-preview})} and {GPT-4 Turbo (\texttt{1106-preview})} through the Microsoft Azure OpenAI API, setting the temperature to 0 for all our experiments. We conducted both authorship verification and attribution experiments three times and calculated the average score. We use py3langid \footnote{https://github.com/adbar/py3langid} to filter out non-English texts. For running the quantized versions of Llama 2 (Llama-2-70B-chat-GPTQ) \cite{frantar2023gptq} and Mistral (Mistral-7B-Instruct-v0.2) \cite{jiang2023mistral}, we utilize an NVIDIA RTX A6000 with 48 GB of GPU memory. Both models are configured with the temperature set to 0 and top\_p set to 1.

\begin{table}[h]
\centering
\resizebox{0.99\linewidth}{!}{
\begin{tabular}{llllll}
\toprule
Dataset                & Name   & Accuracy & Precision & Recall & F1     \\
\midrule
\multirow{2}{*}{Blog}  & TF-IDF & 53.33    & 100.00    & 6.67   & 12.50  \\
                       & BERT   & 50.00    & 50.00     & 100.00 & 66.67  \\
\midrule                       
\multirow{2}{*}{Email} & TF-IDF & 73.33    & 100.00    & 46.67  & 63.64  \\
                       & BERT   & 50.00    & 50.00     & 100.00 & 66.67  \\
\bottomrule
\end{tabular}
}
\caption{Authorship Verification results on the Blog and the Email Dataset for BERT and TF-IDF.}
\label{table:verify-bert}
\end{table}

\section{Additional Results} 
\label{sec:appendix-add-res}
A challenge in evaluating zero-shot authorship verification is comparing our approach with conventional models, which often rely on trained classifiers for classification tasks. To ensure a fair comparison, we adapt these models to fit within a zero-shot framework. To establish a comparison, we consider null accuracy, which is 50\% in a perfectly balanced dataset. Our experiments suggest that using cosine similarity scores of BERT embeddings are mostly distributed around 0.9 We use a threshold of 0.5, where above 0.5 means the same authorship, and vice versa. 

The results shown in Table \ref{table:verify-bert} mean that the BERT model exhibits a tendency to classify each pair of texts as having been authored by the same individual, resulting in a notably high recall rate. In contrast, the TF-IDF approach is characterized by high precision paired with low recall. This indicates that the model predominantly identifies pairs as being written by different authors.

\section{Scientific Artifacts}
We use open-source scientific artifacts in this work, including pandas \cite{mckinney-proc-scipy-2010}, pytorch \cite{paszke2019pytorch}, HuggingFace transformers \cite{wolf2020transformers}, sklearn \citep{scikit-learn}, and NumPy \cite{harris2020arraynumpy}.

\section{Prompt Design} 
\label{sec:appendix-prompt}
This section provides details about the prompt we used for authorship verification (Figure \ref{fig:prompt_verify}) and attribution tasks (Figure \ref{fig:prompt_aa}). Including the system and user instructions for four levels of prompt designs including "{LIP}", "\texttt{no\_guidance}", "\texttt{little\_guidance}", and "\texttt{grammar\_guidance}".

\begin{figure*}[!t]

    \begin{tcolorbox}[colback=promptcolor!10!white,colframe=gray!75!black,title=\text{\textsc{Authorship Verification}:}]
    \textbf{System instruction}: Respond with a JSON object including two key elements:\\
    "analysis": Reasoning behind your answer.\\
    "answer": A boolean (True/False) answer.

    \tcblower
    
        \textbf{Prompting with no guidance}: Verify if two input texts were written by the same author. Input text 1: <text 1>, text 2: <text 2>\\\\
        \textbf{Prompting with style guidance}: Verify if two input texts were written by the same author. Analyze the writing styles of the input texts, disregarding the differences in topic and content. Input text 1: <text 1>, text 2: <text 2>\\\\
        \textbf{Prompting with grammar guidance}: Verify if two input texts were written by the same author. Focus on grammatical styles indicative of authorship. Input text 1: <text 1>, text 2: <text 2>\\\\
        \textbf{Linguistically Informed Prompting (LIP)}: Verify if two input texts were written by the same author.  Analyze the writing styles of the input texts, disregarding the differences in topic and content. Reasoning based on linguistic features such as phrasal verbs, modal verbs, punctuation, rare words, affixes, quantities, humor, sarcasm, typographical errors, and misspellings. Input text 1: <text 1>, text 2: <text 2>
    
    \end{tcolorbox}
    
    \caption{\textbf{Prompt Design for the Authorship Verification Task}.}
    \label{fig:prompt_verify}
\end{figure*}

\begin{figure*}[!t]
\begin{tcolorbox}
[colback=promptcolor!10!white,colframe=gray!75!black,title=\text{\textsc{Authorship Attribution}:}]
    \textbf{System instruction}: Respond with a JSON object including two key elements:\\
    "analysis": Reasoning behind your answer.\\
    "answer": The query text's author ID.

    \tcblower
    
        \textbf{Prompting with no guidance}: Given a set of texts with known authors and a query text, determine the author of the query text. Input query text: <query text>; Texts from potential authors: <candidate texts>\\\\
        \textbf{Prompting with style guidance}: Given a set of texts with known authors and a query text, determine the author of the query text. Do not consider topic differences. Input query text: <query text>; Texts from potential authors: <candidate texts>\\\\
        \textbf{Prompting with grammar guidance}: Given a set of texts with known authors and a query text, determine the author of the query text. Focus on grammatical styles. Input query text: <query text>; Texts from potential authors: <candidate texts>\\\\
        \textbf{\textbf{Linguistically Informed Prompting (LIP)}}: Given a set of texts with known authors and a query text, determine the author of the query text. Analyze the writing styles of the input texts, disregarding the differences in topic and content. Focus on linguistic features such as phrasal verbs, modal verbs, punctuation, rare words, affixes, quantities, humor, sarcasm, typographical errors, and misspellings. Input query text: <query text>; Texts from potential authors: <candidate texts>\\

    \end{tcolorbox}
    
    \caption{\textbf{Prompt Design for the Authorship Attribution Task}. ``query text'' is the text whose authorship needs to be identified. ``candidate texts'' are a collection of texts written by each potential author, which is a JSON object formatted with author IDs as keys and values containing the texts written by them. }
    \label{fig:prompt_aa}
\end{figure*}

\clearpage

\end{document}